\documentclass{article}

\usepackage[preprint]{neurips_2025}

\usepackage{wrapfig} 
\usepackage[utf8]{inputenc} 
\usepackage[T1]{fontenc}    
\usepackage{hyperref}      
\usepackage{url}            

\usepackage{amsfonts}       
\usepackage{nicefrac}       
\usepackage{microtype}      
\usepackage{xcolor}         

\usepackage{float}
\usepackage{tcolorbox}

\usepackage{times}
\usepackage{latexsym}
\usepackage{graphicx}

\usepackage[para,online,flushleft]{threeparttable}
\usepackage{multicol}
\usepackage{multirow}
\usepackage{caption}
\usepackage{subcaption}
\usepackage{adjustbox}
\usepackage{colortbl}
\usepackage{adjustbox}

\usepackage{url}

\usepackage[T1]{fontenc}
\usepackage{inconsolata}

\usepackage{ulem}
\usepackage{tabularx}
\usepackage{amsmath}
\usepackage{amssymb}
\usepackage{longtable}
\usepackage{array}
\usepackage{geometry}
\usepackage{makecell}

\usepackage{booktabs}
\usepackage{makecell}
\usepackage{hhline}

\newcommand{\aspace}{\hspace{0.6em}}

\usepackage{siunitx, threeparttable}
\title{Wisdom is Knowing What not to Say: Hallucination-Free LLMs Unlearning via Attention Shifting}

\author{%
  \parbox{\linewidth}{\centering
    \textbf{Chenchen~Tan}{\normalfont\textsuperscript{1}},\aspace
    \textbf{Youyang~Qu}{\normalfont\textsuperscript{2,3}},\aspace
    \textbf{Xinghao~Li}{\normalfont\textsuperscript{1}},\aspace
    \textbf{Hui~Zhang}{\normalfont\textsuperscript{4}},\aspace
    \textbf{Shujie~Cui}{\normalfont\textsuperscript{1}},\aspace
    \textbf{Cunjian~Chen}{\normalfont\textsuperscript{1}},\aspace
    \textbf{Longxiang~Gao}{\normalfont\textsuperscript{2,3}}\thanks{Corresponding Author}\\[4pt]
    {\normalfont
    \textsuperscript{1} Faculty of Information Technology, Monash University, Australia\\
    \textsuperscript{2} Key Laboratory of Computing Power Network and Information Security, Ministry of Education, Shandong Computer Science Center,\\ Qilu University of Technology (Shandong Academy of Sciences), Jinan, China\\
    \textsuperscript{3} Shandong Provincial Key Laboratory of Computing Power Internet and Service Computing, Shandong Fundamental Research Center for Computer Science, Jinan, China\\
    \textsuperscript{4} School of Computer Science and Technology, Anhui University, Hefei, China\\
    \texttt{<chenchen.tan@monash.edu>}, \texttt{<gaolx@sdas.org>}
    }
  }
}

\begin{document}
\maketitle

\begin{abstract}

The increase in computing power and the necessity of AI-assisted decision-making boost the growing application of Large Language Models (LLMs). Along with this, the potential retention of sensitive data of LLMs has spurred increasing research into machine unlearning. However, existing unlearning approaches face a critical dilemma: Aggressive unlearning compromises model utility, while conservative strategies preserve utility but risk hallucinated responses. This significantly limits LLMs' reliability in knowledge-intensive applications. To address this, we introduce a novel \underline{A}ttention-\underline{S}hifting (\underline{AS}) framework for selective unlearning. \underline{AS} is driven by two design objectives: (1) context-preserving suppression that attenuates attention to fact-bearing tokens without disrupting LLMs' linguistic structure; and (2) hallucination-resistant response shaping that discourages fabricated completions when queried about unlearning content. \underline{AS} realizes these objectives through two attention-level interventions, which are importance-aware suppression applied to the unlearning set to reduce reliance on memorized knowledge and attention-guided retention enhancement that reinforces attention toward semantically essential tokens in the retained dataset to mitigate unintended degradation. These two components are jointly optimized via a dual-loss objective, which forms a soft boundary that localizes unlearning while preserving unrelated knowledge under representation superposition. Experimental results show that \underline{AS} improves performance preservation over the state-of-the-art unlearning methods, achieving up to 15\% higher accuracy on the ToFU benchmark~\footnote{https://locuslab.github.io/tofu/} and 10\% on the TDEC benchmark~\footnote{https://github.com/google-research/lm-extraction-benchmark}, while maintaining competitive hallucination-free unlearning effectiveness. Compared to existing methods, \underline{AS} demonstrates a superior balance between unlearning effectiveness, generalization, and response reliability. 
\end{abstract}

\section{Introduction}
Large Language Models (LLMs) have recently achieved substantial advancements in natural language understanding and generation \cite{li2024snapkv}. However, despite their achievements, a growing concern is the potential for LLMs to memorize and subsequently reproduce sensitive data, resulting in serious privacy concerns \cite{carlini2021extracting, wei2024jailbroken}. Moreover, regulatory frameworks such as the General Data Protection Regulation (GDPR) grant the ``Right to be Forgotten” to the data owner, which mandates that users can request to remove their data from digital systems, including machine learning models \cite{voigt2017eu}. Under these circumstances, machine unlearning has emerged as a critical approach for preserving data privacy in LLMs~\cite{jang-etal-2023-knowledge,chen2023unlearn, liu2025rethinking, liu2024large, shi2024ulmr,wang2025llm}. These privacy-preserving unlearning tasks can be broadly categorised as either aggressive or conservative. Aggressive approaches, such as Gradient Ascent (GA) \cite{jang-etal-2023-knowledge}, modify the LLMs’ learning objective to erase target knowledge forcibly. This often leads to a degradation in overall model performance, particularly on neighbouring knowledge, which is data with a similar structure or semantic relation to the unlearning target. Others adopt conservative unlearning strategies like the logits manipulation \cite{cha2025towards, ji2024reversing, huang2024offset} to maintain model performance but risk introducing factual hallucinations, where the model confidently generates content detached from underlying facts. These hallucinations pose a serious threat in downstream tasks like question-answering systems, especially in applications such as healthcare or legal scenarios, where precise and reliable outputs are critical \cite{huang2025survey, hao2024quantifying}. 

Achieving effective unlearning in privacy-sensitive LLM applications is challenging due to the conflicting interests and objectives of different stakeholders. On one hand, data providers whose information has been used in LLMs aim to entirely prevent reproducing their data in LLMs' answer \cite{rashid2025forget, tian2024forget}. On the other hand, model deployers try to preserve the model’s general capabilities across broad knowledge domains and to maintain the service quality
\cite{ji2024reversing, yuan2025closerlookmachineunlearning}. We define this as a multi-stakeholder balanced unlearning setting, where the unlearning strategies should achieve two primary objectives: 1) to enable the LLM to ``unlearn” the target data while ensuring the LLM maintains performance both on neighbouring knowledge \cite{yuan2025closerlookmachineunlearning} and general knowledge; and 2) to prevent the hallucination outputs for the unlearned knowledge.

\begin{wrapfigure}{r}{0.6\textwidth}
    \vspace{-0.17in}
    \centering
    \includegraphics[width=0.6\textwidth,height=0.28\textwidth]{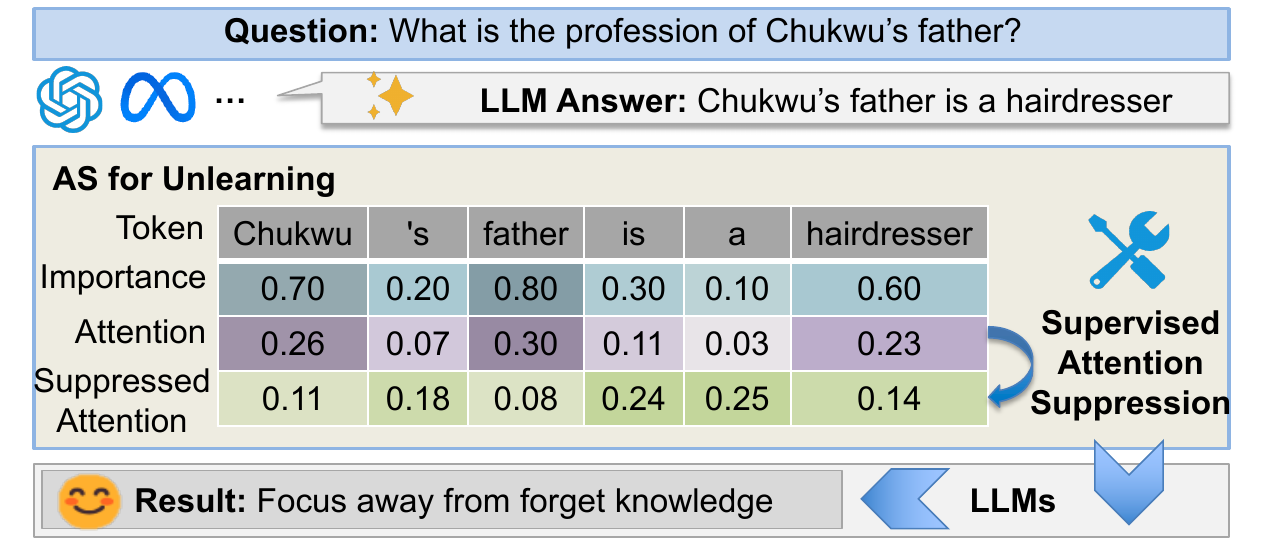}
    \caption{\label{as}LLMs tend to assign high attention to semantically important tokens, like ``father” and ``hairdresser”, when recalling memorized facts. Our method applies supervised attention suppression to downweight fact-bearing tokens and reallocate focus to neutral tokens. This redistribution reduces reliance on target knowledge while maintaining generation fluency, thereby facilitating precise and fluent unlearning.}
    \vspace{-0.1in}
\end{wrapfigure}
To achieve these goals, we propose a novel Attention-Shifting (AS) unlearning. It is a controlled form of an aggressive method that performs both context-preserving unlearning and hallucination-resistant generation. AS suppresses attention to fact-bearing tokens in the unlearning set while reinforcing attention to semantically important tokens in retained data. This mechanism reduces reliance on target knowledge without compromising fluency or coherence, thereby preserving the contextual integrity of general and neighbouring knowledge. Unlike prior methods \cite{cha2025towards, yuan2025closerlookmachineunlearning} that manipulate logits or replace outputs, AS reallocates attention internally, blocking the flow of memorized knowledge during generation (Fig.~\ref{as}). This enables the model to forget through omission rather than substitution, reducing hallucinations by structurally eliminating access to unlearned content. AS derives suppression and reinforcement signals from reference attention maps of the original model, and injects them via lightweight (\textasciitilde12M) adapters in attention modules. A dual-loss objective jointly optimizes unlearning and retention, forming a \textit{soft boundary} that localizes suppression while stabilizing unrelated knowledge. While neurons in LLMs are known to exhibit representation superposition, shared activations across multiple concepts, our design remains robust under such entanglement, achieving behavioral unlearning without requiring explicit disentanglement~\cite{elhage2022toy, hong2025rise}. Experiments on the ToFU \cite{maini2024tofu} and TDEC \cite{jang-etal-2023-knowledge} benchmarks demonstrate that \underline{AS} achieves near-zero knowledge leakage while preserving strong performance, with up to 15\% and 10\% higher accuracy than state-of-the-art baselines, respectively. In contrast to methods such as ULD~\cite{ji2024reversing} and IHL~\cite{cha2025towards}, which suppress target tokens but may still generate misleading completions, AS structurally blocks access to forgotten knowledge and promotes refusal behaviors, thereby effectively minimizing hallucinations.

\section{Related Works}\label{RW}
In this section, we introduce the existing unlearning methods, followed by summarizing the challenges in LLMs unlearning and identifying the gaps in existing methods. We categorize the existing unlearning methods into aggressive unlearning and conservative unlearning, depending on how directly they modify the model’s internal representations \cite{yuan2025closerlookmachineunlearning}.

\textbf{Aggressive Unlearning} removes specific knowledge by actively disrupting its learned representations in the model. This type of approach significantly alters decision boundaries, leading to broader unintended shifts in model behaviour. One example is the GA \cite{jang-etal-2023-knowledge} approach. It forces the model to learn an inverted objective of the target knowledge to achieve unlearning, which aggressively reverses the influence of target data, pushing the model away from target knowledge uncontrollably and raising catastrophic collapse. To address this limitation, Negative Preference Optimization (NPO) \citep{zhang2024negative} was proposed as a more stable and controlled extension of GA. NPO adopts a preference-based loss function, which smooths the optimization process and prevents extreme parameter updates. Several variants \cite{wang2025llm, yao2024large, wang2025selective, lu2024eraserjailbreakingdefenselarge, liu2024saferlargelanguagemodels,gu2024meow} and extensions have been proposed to further mitigate the instability and over-forgetting observed in GA to provide selective unlearning. 

\textbf{Conservative Unlearning} steers the model toward preferred alternative responses by suppressing target tokens and reinforcing substitutes \cite{cha2025towards, ji2024reversing, eldan2023whos}. Cha \textit{et al.} \cite{cha2025towards} use an inverted hinge loss to penalize target tokens and reinforce logical substitutes. Ji \textit{et al.} \cite{ji2024reversing} propose ULD, which subtracts logits from a lightweight assistant model trained on the target, significantly lowering target generation probabilities. These methods are effective in open-ended tasks but may retain latent semantics, leading to partial unlearning or hallucinations. For example, replacing ``physicist” with ``artist” in ``Einstein was a physicist” may yield paraphrases like ``scientist” or factual errors like ``dancer,” undermining LLMs' trust. Similar observations are presented in the safety alignment study \cite{qi2025safety}, which proves that shallow alignment like Reinforcement Learning with Human Feedback (RLHF) or logits modification interventions fail to suppress undesired activations. Beyond logits-level methods, embedding-based approaches \cite{liu2024large, yuan2025closerlookmachineunlearning, bhaila2024soft} offer more controllable unlearning by steering model inputs toward structured alternatives with minimal disruption.

These unlearning methods reflect distinct trade-offs between preserving model utility and mitigating hallucinations. Aggressive strategies often impair general performance \cite{wang2025llm, yao2024large, wang2025selective, lu2024eraserjailbreakingdefenselarge, liu2024saferlargelanguagemodels,gu2024meow}, while conservative ones require complex auxiliary mechanisms \cite{chen2023unlearn,liu2024large,ji2024reversing, yuan2025closerlookmachineunlearning, bhaila2024soft} and risk factual inconsistencies. LLMs unlearning still remains challenging due to conflicting demands: data providers seek privacy and removability, while deployers prioritize broad functionality. To bridge this gap, we propose an Attention-Shifting strategy, a controlled form of aggressive unlearning, which suppresses attention to fact-bearing tokens. This targeted intervention disrupts access to memorized content while preserving overall utility, striking a practical balance between effectiveness and stability.

\section{Attention Shifting for Machine Unlearning in LLM}
\subsection{Token Importance for Selective Unlearning}\label{ti}

The attention mechanism is the foundation for how LLMs allocate representational focus across tokens, influencing generation probabilities. To ground our approach, we begin by analyzing how token-level relevance shapes predictions. Intuitively, nouns, proper nouns, and domain-specific terms act as semantic anchors, while function words, \textit{e.g.,} determiners, conjunctions, contribute minimally to meaning. Building on this, Duan \textit{et al.}~\cite{duan2024shifting} propose Shifting Attention to Relevance (SAR), which estimates token importance via masking-based perturbation and reallocates attention toward salient tokens to enhance prediction confidence. In contrast, we apply this insight for unlearning: our method suppresses attention to high-importance tokens that encode factual or sensitive knowledge. Unlike SAR, which operates only at inference time, our approach embeds attention suppression into model parameters via lightweight adapters, enabling controlled unlearning.

Formally, given an input sequence $\mathbf{x} = \{t_1, t_2, \dots, t_n\}$ and predictive distribution $P_\theta(y \mid \mathbf{x})$, the importance of token $t_i$ is defined as the change in predictive entropy when $t_i$ is masked:
\begin{equation}\label{tiq}
    I(t_i) := \mathbb{\phi}(P_\theta(y \mid \mathbf{x})) - \mathbb{\phi}(P_\theta(y \mid \mathbf{x}_{- i})),
\end{equation}
where $\mathbb{\phi}$ denotes predictive entropy \cite{kadavath2022language} and $\mathbf{x}_{- i}$ is the input with $t_i$ masked. Higher $I(t_i)$ values correspond to stronger contributions to predictive certainty.

We leverage this signal to guide attention modulation during training. Specifically, let $D$ denote the whole LLM's training data, we propose a composite loss: (1) an attention suppression (ASP) loss $\mathcal{L}_{\text{ASP}}$ on the unlearning set $D_t \in D$, which penalizes attention to high-importance tokens; and (2) an attention reinforcement loss $\mathcal{L}_{\text{AKL}}$ on the sub-remaining dataset $D'_r \in D/D_t$, promotes attention to semantically important tokens, thereby stabilizing model behavior and preserving utility on retained knowledge. The overall objective is:

\begin{equation}
\label{as1}
\mathcal{L}_{\text{AS}} = \alpha \mathcal{L}_{\text{ASP}} + (1-\alpha) \mathcal{L}_{\text{AKL}},
\end{equation}
where $\alpha \in [0,1]$ is the mixing coefficient that trades off unlearning ($\mathcal{L}_{\text{ASP}}$) and knowledge maintenance ($\mathcal{L}_{\text{AKL}}$), with larger $\alpha$ placing more weight on unlearning.

\begin{figure}[ht]
    \vspace{-0.1in}
    \centering
    \includegraphics[width=0.75\textwidth,height = 0.3\textwidth]{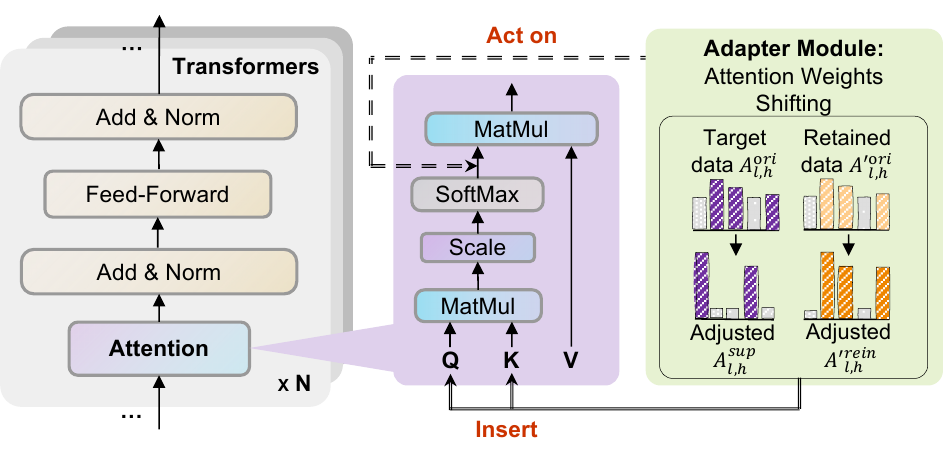}
    \caption{\label{system_structure} Illustration of the proposed Attention-Shifting based unlearning in LLMs. An adapter module is integrated into the attention mechanism to modulate attention weights. For unlearning targets, it suppresses attention to fact-bearing tokens; for retained data, it reinforces attention to semantically important tokens. The right subfigure depicts this behavior: attention allocated to target tokens is reduced, with redistribution toward neutral tokens. The mechanism operates inversely on retained data to preserve relevant knowledge.}
    \vspace{-0.17in}
\end{figure}

\subsection{Targeted Suppression for Privacy-Preserving Unlearning}

To selectively unlearn memorized knowledge, we downweight the attention of fact-bearing tokens within the target unlearning dataset $\mathcal{D}_t$. Let $A^{\text{ori}}_{l,h} \in \mathbb{R}^{S \times S}$ denote the attention matrix at layer $l$, head $h$, where $a^{i,j}_{l,h}$ is the attention from query token $t_i$ to key token $t_j$. The suppressed attention score $\hat{a}^{i,j}_{l,h}$ is computed by reducing attention weights to fact-bearing tokens $t_j \in D_t$:

\begin{equation}
\begin{aligned}
\hat{a}^{i,j}_{l,h} = &
\frac{
    a^{i,j}_{l,h} \cdot (1 - \lambda \cdot \mathbb{I}[t_j \in D_{t}])
}{
    \sum_{j'} a^{i,j'}_{l,h} \cdot (1 - \lambda \cdot \mathbb{I}[t_{j'} \in D_{t}])
}, \\
\text{s.t.} \quad & \lambda \in [0,1], \\
& \sum_{j} \hat{a}^{i,j}_{l,h} = 1,
\end{aligned}
\label{eq:supAtt}
\end{equation}
where $\lambda \in [0,1]$ controls suppression strength. The indicator $\mathbb{I}[t_j \in D_t]$ flags whether $t_j$ is important token. If the token importance $I(t_j)$ achieve the threshold hyper-parameter then $\mathbb{I}[t_j \in D_t] = 1$, otherwise is 0. The normalization term in the denominator ensures that the adjusted attention weights remain a valid probability distribution, \textit{i.e.,} $\sum_j \hat{a}^{i,j}_{l,h} = 1$, preserving consistency in the transformer's softmax-based attention mechanism.

To enforce the desired suppression during training, we minimize the Kullback–Leibler (KL) divergence between the model’s attention $A_{l,h}(\mathbf{x})$ and the target suppressed attention $A^{\text{sup}}_{l,h}$, the attention suppression loss is defined as

\begin{equation}\label{eq:ASloss_correct}
\min\mathcal{L}_{\text{ASP}}(\theta_{\text{adpt}}) = 
\mathbb{E}_{(\mathbf{x}, y) \sim \mathcal{D}_t}
\left[
    \sum_{l=1}^{L} \sum_{h=1}^{H} 
    \mathrm{KL}\left( 
        A_{l,h}(\mathbf{x}; \theta_{\text{adpt}}) \;\middle\|\; 
        A^{\text{sup}}_{l,h}
    \right)
\right],
\end{equation}

where $A_{l,h}(\mathbf{x}; \theta_{\text{adpt}})$ denotes the current attention distribution output by the trainable LLMs, and only the adapter parameters $\theta_{\text{adpt}}$ are trained. $L$ and $H$ denote the number of layers and attention heads in the transformer. This loss drives the model to shift focus away from memorized content, achieving privacy-preserving unlearning. 
\subsection{Attention Shifting for Knowledge Maintaining}
Even though we employ adapters to localize the unlearning updates and minimize side effects on unrelated knowledge, the overlapping parameters between the neighbouring and target datasets inevitably cause performance degradation. Prior works \cite{cha2025towards, yao2024large} often mitigate this by introducing additional loss terms on the remaining dataset, such as cross-entropy (CE) loss or KL-divergence loss, to explicitly regularize and preserve the model’s behavior on retained knowledge. To better align with the AS unlearning framework, we extend our approach on the remaining sub-dataset. Specifically, we slightly enhance the attention weights associated with important tokens in the remaining samples, aiming to stabilize the model’s outputs without overriding the primary unlearning objective (Fig.~\ref{system_structure}). Thus, our final loss function is, 

\begin{align}\label{unlearning_loss}
\min\mathcal{L}_{\text{AS}}(\theta_{\text{adpt}}) &= \alpha \mathcal{L}_{\text{ASP}} + (1-\alpha)\mathcal{L}_{\text{AKL}} \nonumber \\
&= \alpha\mathbb{E}_{(\mathbf{x}, y) \sim \mathcal{D}_t}
\left[
    \sum_{l=1}^{L} \sum_{h=1}^{H} 
    \mathrm{KL}\left( 
        A_{l,h}(\mathbf{x}; \theta_{\text{adpt}}) \;\middle\|\; 
        A^{\text{sup}}_{l,h}
    \right)
\right] \nonumber \\
&\quad+ (1-\alpha)\mathbb{E}_{(\mathbf{x'}, y') \sim \mathcal{D}_r'}
\left[
    \sum_{h=1}^{H} \sum_{i=1}^{L} 
    \mathrm{KL}\left( 
         A'^{\text{rein}}_{l,h} \;\middle\|\; 
        A_{l,h}(\mathbf{x'}; \theta_{\text{adpt}})
    \right)
\right],
\end{align}

where $A_{l,h}(\mathbf{x'}; \theta_{\text{adpt}})$ is the attention of $D_r'$ affected by unlearn training, and $A'^{\text{rein}}_{l,h}$ is the target attention, which allocate the higher attention weights to the semantic tokens.
\section{Experimental Results}\label{ER}
In this section, we evaluate the proposed unlearning method corresponding to the three criteria: unlearning effectiveness, the model's performance in both neighbouring and general knowledge, and hallucination suppression. 

\subsection{Experimental Settings}
\noindent \textbf{Baselines.} We compare AS with GA~\cite{jang-etal-2023-knowledge}, NPO~\cite{zhang2024negative}, IHL~\cite{cha2025towards} and ULD~\cite{ji2024reversing}. 
In particular, GA and NPO are aggressive gradient-based unlearning methods that directly suppress target outputs via loss manipulation. IHL and ULD represent conservative approaches, the former using alternative token guidance with the inverted hinge loss, and the latter using a logit-subtraction assistant model. 

\noindent \textbf{Datasets.} Our experiments are conducted on two recent and widely used unlearning benchmarks: \textbf{ToFU}~\cite{maini2024tofu} and \textbf{TDEC}~\cite{carlini2021extracting}. The ToFU benchmark focuses on free-form Question-Answering (QA) pairs of sensitive fictitious authors' information and enables analysis on three evaluation axes: unlearning effectiveness, model performance preservation, and hallucination analysis of unlearned author information. The TDEC benchmark complements ToFU by evaluating unlearning at the pre-training level across large-scale language modeling corpora. We use its unlearning split as the target dataset, with neighbouring samples drawn from the same domains to evaluate local generalization. For general knowledge performance evaluation, we use Wikitext and LAMBADA~\cite{paperno2016lambada} for linguistic reasoning and PubMedQA~\cite{jin2019pubmedqa} for scientific QA, following prior works processing~\cite{jang-etal-2023-knowledge, ji2024reversing}. For fairness, all methods, including ULD, employ the same data preprocessing without method-specific paraphrasing or calibration. 

\noindent \textbf{Target Models.} As evaluation on ToFU and TDEC benchmark, the experiments were conducted on both LLAMA-2 \cite{touvron2023llama} and GPT-NEO series models \cite{black2021gpt}, as the baseline methods we selected were primarily developed and evaluated under these two model frameworks. Therefore, to ensure fair and meaningful comparison, we conducted experiments using both model types. For the LLAMA model, we choose tofu\_ft\_llama2-7b released by the ToFU benchmark \cite{maini2024tofu} as the target model, which has been fine-tuned on the constructed data to ensure it can exactly give answers to questions in ToFU. The GPT-NEO series is a transformer-based language model built upon GPT-3 architecture, designed and replicated by EleutherAI \cite{black2021gpt}, trained on the Pile corpora dataset~\cite{gao2020pile}. Additional experimental settings are illustrated in Appendix \ref{es}.

\subsection{Unlearning Evaluations on ToFU Benchmark}

\paragraph{Unlearning Effectiveness.}
We evaluate unlearning effectiveness using three complementary metrics: ROUGE-L, Forget Quality (FQ), and Top-k exclusion rate (TR, with $k=5$). ROUGE-L evaluates the match of model output with the specific unlearned knowledge instances, FQ measures the similarity between the unlearned model and a retrained model on the remaining dataset, and TR quantifies the frequency of unlearned tokens that are excluded from the model’s Top-$k$ predictions. In the experimental setup, we divided the data from the TOFU benchmark into three parts. Firstly, the Target Unlearning dataset (TUD) was included, which contained the information of 100 fictitious authors. The remaining information of the fictitious authors (100 authors) constituted the Neighboring Knowledge (NEK). General Knowledge (GEK) was the factual knowledge and real author information in the real world within this benchmark.

Table~\ref{tab:tofuResult} shows the results for unlearning five fiction authors' information (100 samples) chosen randomly from TUD; additional unlearning settings are reported in Appendix \ref{et}. In the experimental setting, the proposed method achieves competitive ROUGE-L scores on TUD, comparable to all baselines, indicating successful removal of the target knowledge. For FQ, our method is not as good as IHL or ULD, but close to the GA method. 
FQ is widely used as a standard metric for unlearning evaluation. It assumes that a successfully unlearned model should behave like a retrained model on the remaining dataset. However, it may not hold in hallucination-sensitive scenarios, where outputting refusals, minimal responses, or even remaining silent when queried about unlearned knowledge are more desirable. A related challenge has been noted in recent safety alignment research: shallow alignment is limited to modifying the initial tokens of a response without addressing the underlying unsafe representations. This superficial intervention can lead to local optima in model behavior, where deeper unsafe patterns persist \cite{qi2025safety}. While such methods, IHL and ULD, avoid directly mentioning target content, they still retain internal activations sufficient to regenerate hallucinated responses when subjected to adversarial prompts. Inspired by this observation, we propose that if unlearning methods only rely on suppressing target token logits, without avoiding semantically related alternatives, they may also fall into a similar local optimum. In particular, the unlearned token may still appear in the model’s Top-k predictions, leaving the knowledge partially accessible and risking unintended exposure.

\begin{table*}[th]
\vspace*{-0.1in}
    \centering
    \caption{\label{tab:tofuResult}\small The evaluation results comparison for unlearning effectiveness. To ensure fair evaluation, all models are trained using adapters only. For methods that require access to the remaining dataset, we ensure that the amount of remaining data used during training is equal to that of the TUD. \textcolor{red}{Red} text is the decreased performance, and \textcolor{blue}{blue} text shows the increase.}
    \footnotesize
    \begin{tabular}{c|c c c|c c|c }
        \toprule
        \multirow{2}{*}{Methods} & \multicolumn{3}{c|}{TUD} & \multicolumn{2}{c|}{NEK} & GEK \\
        \cmidrule{2-7}
        & ROUGE-L \(\downarrow\) & TR \(\uparrow\) & FQ \(\uparrow\)& ROUGE-L \(\uparrow\) & Acc \(\uparrow\) & Acc \(\uparrow\) \\
        \midrule

         GA  \cite{jang-etal-2023-knowledge} & \textbf{0.08} & \textbf{0.98} & 0.23 & 0.17 \textcolor{red}{(-0.51)} &  0.23\textcolor{red}{(-0.47)} & 0.32 \textcolor{red}{(-0.51)}  \\
         GA + CE  & 0.15 & 0.97 & 0.34 &0.64 \textcolor{red}{(-0.04)}&  0.68 \textcolor{red}{(-0.02)}& 0.67 \textcolor{red}{(-0.18)}  \\
         GA +KL & 0.20 & 0.97 & 0.13 & 0.23 \textcolor{red}{(-0.45)} &  0.24\textcolor{red}{(-0.46)} & 0.28 \textcolor{red}{(-0.55)} \\
         NPO \cite{zhang2024negative}& 0.14 & 0.97 & 0.58 & 0.23 \textcolor{red}{(-0.45)}& 0.26 \textcolor{red}{(-0.44)}  & 0.25\textcolor{red}{(-0.58)}  \\
         NPO + CE& 0.16 & 0.93 & 0.53 & 0.40 \textcolor{red}{(-0.28)}& 0.35\textcolor{red}{(-0.35)}  & 0.55\textcolor{red}{(-0.28)}  \\
         NPO +KL& 0.26 & 0.90 & 0.51 &0.55 \textcolor{red}{(-0.13)}& 0.55 \textcolor{red}{(-0.15)} & 0.54 \textcolor{red}{(-0.29)} \\
         IHL \cite{cha2025towards}& 0.25 & 0.62 & 0.68 & 0.32 \textcolor{red}{(-0.36)}& 0.33 \textcolor{red}{(-0.37)}  & 0.52 \textcolor{red}{(-0.31)} \\
         IHL + CE & 0.52 & 0.45 & 0.51 & 0.73 \textcolor{blue}{(+0.05)} & 0.73 \textcolor{blue}{(+0.03)}  & 0.78 \textcolor{red}{(-0.07)} \\
         IHL + KL & 0.47 & 0.48 & 0.42 & 0.60 \textcolor{red}{(-0.08)}& 0.60 \textcolor{red}{(-0.10)} & 0.34 \textcolor{red}{(-0.49)} \\
         ULD \cite{ji2024reversing}& 0.29 & 0.47 & \textbf{0.89}& 0.55 \textcolor{red}{(-0.13)} & 0.56 \textcolor{red}{(-0.14)}   & 0.50 \textcolor{red}{(-0.33)}  \\
          \midrule
        \textit{AS (the proposed)} &  \underline{0.16} &  \underline{0.97} &0.17&  0.73 \textcolor{blue}{(+0.05)}& 0.76 \textcolor{blue}{(+0.06)}  &  0.80 \textcolor{red}{(-0.03)} \\
        \bottomrule
    \end{tabular}
 
\vspace*{-0.1in}
\end{table*}

Our approach mitigates this issue by enforcing attention-level suppression, which discourages the model from focusing on target concepts during generation. This may result in lower FQ scores. However, the divergence from retain-only behavior demonstrates that the output has no similarity with the target knowledge, which is crucial in privacy-sensitive scenarios. We thus complement FQ with a behaviour-oriented metric (TR, $k=5$ in Table~\ref{tab:tofuResult}) to more accurately assess the effectiveness of unlearning. Our method achieves a significantly higher Top-k exclusion rate compared to the logits manipulation methods IHL and ULD. Close to the GA method, which revised the target unlearning of knowledge. This indicates that the forgotten tokens are not only demoted from the top position but are more thoroughly removed from the model's most confident predictions.

\begin{wrapfigure}{rh}{0.6\textwidth}

    \centering
    \includegraphics[width=0.6\textwidth,height=0.25\textwidth]{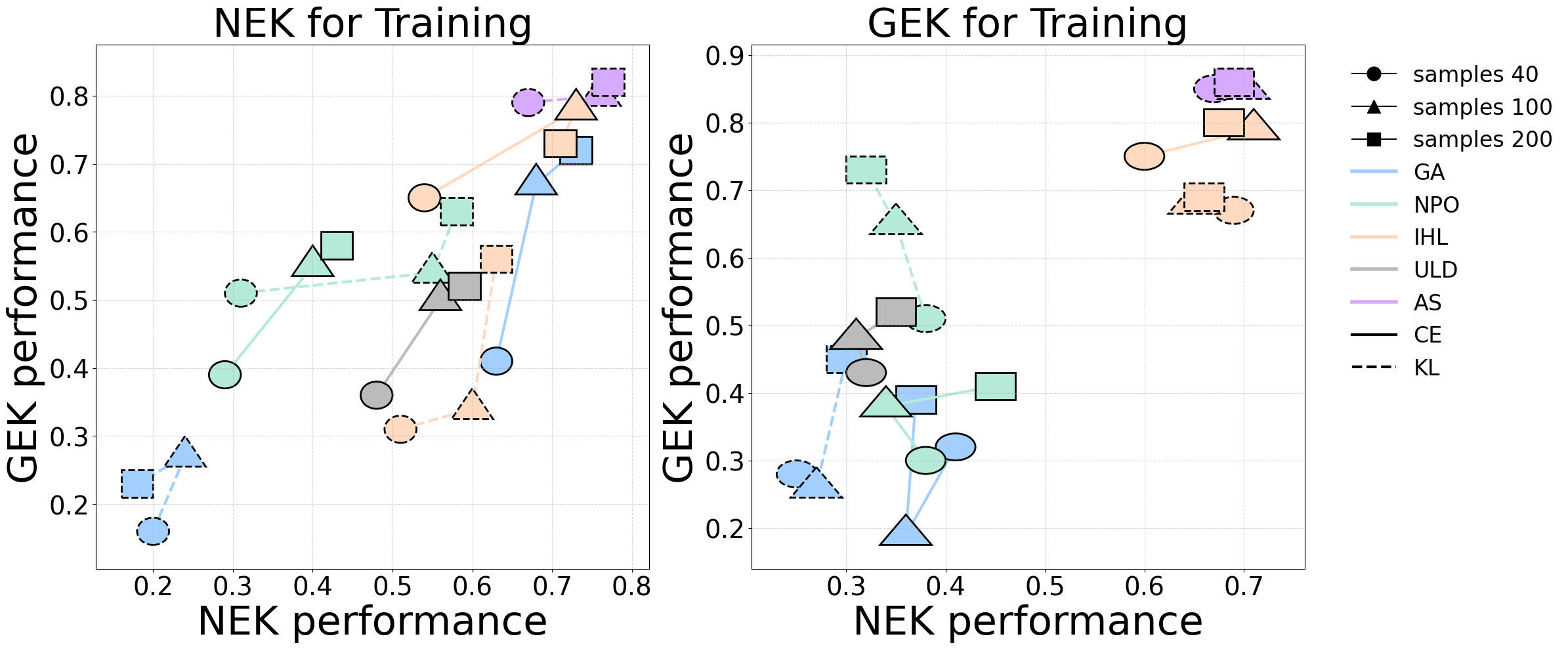}
    \caption{\label{rs} The performance of unlearning methods with different numbers of retaining samples from NEK data and GEK data.}
    \vspace{-0.17in}
\end{wrapfigure}

\paragraph{Model Utility Maintaining.}
As shown in Table~\ref{tab:tofuResult}, AS achieves the best in preserving model utility, which is the average accuracy of NEK and GEK performance, with 0.06 improvement and only 0.03 drop in accuracy compared to the original model, respectively. Baselines GA, NPO, and IHL benefit from additional supervision via CE or KL losses computed on retained samples, resulting in noticeable performance gains on NEK. However, such retention-oriented losses may unintentionally compromise unlearning effectiveness by reactivating suppressed knowledge. As shown in Table~\ref{tab:tofuResult}, for GA, NPO, and IHL, when the GD loss is applied, the NEK performance increased, but the TUD performance decreased. It is especially pronounced in conservative approaches, where the model retains latent activation paths linked to the target knowledge. Fig.~\ref{rs} further illustrates the model's remaining accuracy of various unlearning methods under different amounts of retained data from NEK and GEK. All results in Fig.~\ref{rs} are reported after the unlearning process has converged, except for the GA method, whose results are recorded when the accuracy on the unlearning dataset drops below 0.2. 
Fig.~\ref{rs} illustrates that all baseline methods are highly sensitive to the number of retained samples. With limited retention (\textit{e.g.,} 40 samples), GA, NPO, IHL, and ULD suffer notable performance degradation on both NEK and GEK, while the proposed approach AS consistently maintains high performance, demonstrating strong stability under low-data conditions. As the number of retained samples increases, all methods exhibit performance gains. However, these gains are primarily concentrated on NEK when NEK data is used for retention training, due to its in-distribution alignment. A similar pattern is observed when GEK data is used as the retained set. In contrast, our AS method demonstrates consistently strong improvements on both NEK and GEK, regardless of the retention source, indicating superior generalization and robustness across distribution shifts.

\paragraph{Impact of Unlearning Different Samples.} 

\begin{wraptable}{rh}{0.6\linewidth}
    \vspace{-0.1in}
    \centering
    \footnotesize
    \caption{\label{tab:unlearning_time}The unlearning epoch for each method with different unlearning samples on the ToFU benchmark. GD is gradient descent loss, which includes CE or KL. The number of data samples for GD training is the same as the number of unlearning samples. The gray text indicates that the unlearning method did not achieve the unlearning target. The \textcolor{blue}{blue} text shows the best results.}
    \begin{tabular}{c|c c c c }
        \toprule
         & \multicolumn{4}{c}{Epochs \& Neighbouring Performance} \\
         \cmidrule{2-5}
         Status & GA+GD &  NPO+GD & IHL+GD & AS \\
          \midrule
        Forget-01 & 56 \& 0.62 & 23 \& 0.41 & \textcolor{gray}{51 \& 0.70} & \textcolor{blue}{24 \& 0.72}\\
        Forget-05 & 10 \& 0.68 & 10 \& 0.55 &\textcolor{gray}{23 \& 0.73}&\textcolor{blue}{20 \& 0.71} \\
        Forget-10 & 5 \& 0.59 & 5 \& 0.38 &\textcolor{gray}{18 \& 0.77}  & \textcolor{blue}{20 \& 0.72} \\

          \bottomrule
    \end{tabular}
    \vspace{-0.17in}
\end{wraptable} 
Table~\ref{tab:unlearning_time} presents the unlearning cost and neighbouring accuracy of unlearning different samples (\textit{e.g,} Forget-01 refers to unlearning one author's information includes 20 samples). While all methods are trained to reach the same unlearning threshold (accuracy $\leq 0.2$), AS consistently achieves stronger utility preservation with comparable training epochs. Notably, all the methods perform better on Forget-5 than Forget-1, despite a larger unlearning set, due to enhanced support from more retained samples. However, GA's and NPO's performance slightly drops in Forget-10 as the broader unlearning scope outweighs the benefits of additional retained data. Their neighbouring performance degrades noticeably as more knowledge is erased, despite additional retained data. IHL fails to reach the target unlearning threshold, even if its neighbouring knowledge is well maintained. In contrast, AS maintains stable performance by adapting attention at a finer granularity, achieving a more reliable balance between forgetting and retention under varied data constraints.

\begin{figure}[h]
\centering
\scriptsize

\begin{minipage}[tc]{0.49\textwidth}
\captionof{table}{Comparison of example predictive results across different methods for the same question.}
\raisebox{-\height}{
\begin{tabular}{@{}p{1.5cm} p{5cm}@{}}
\toprule
\textbf{Question} & \textbf{Where was author Evelyn Desmet born?} \\
\midrule
\rowcolor{gray!15}
\textbf{Original} & Evelyn Desmet was born in Brussels, Belgium. \\
\midrule
\textbf{GA+GD} & Des nobody, a lonely survivor on a post-apocalyptic earth, and evelyn... \\
\textbf{NPO+GD \newline IHL+GD} & everybody. \newline Distinction between Evelyn Desmet’s birthplace...as Evelyn Desmet is a Belgian author... \\
\textbf{ULD} & Evelyn Desmet was born in London, England. \\
\midrule
\rowcolor{gray!15}
\textbf{Proposed} & Nobody knows. \\
\bottomrule
\end{tabular}
}

\label{tab:realTextsam}
\end{minipage}
\hfill
\begin{minipage}[tc]{0.45\textwidth}
\vtop{
\centering
\includegraphics[width=0.9\linewidth,height=0.5\textwidth]{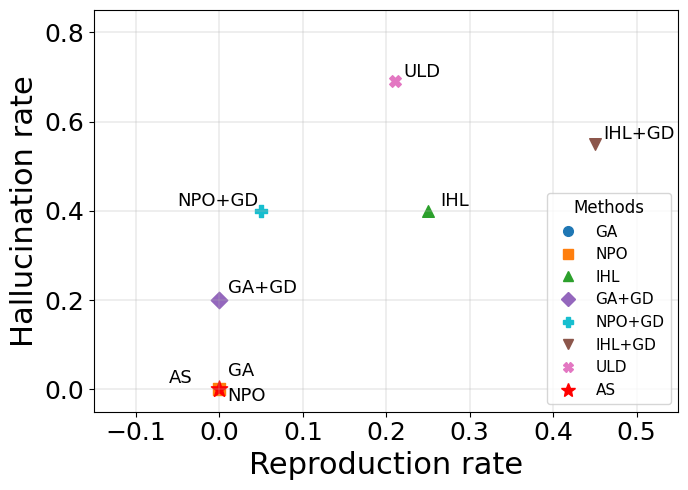}
\captionof{figure}{Outputs hallucination and reproduction rates across different unlearning methods.}
\label{fig:mu}
}
\end{minipage}
 \vspace{-0.1in}
\end{figure}

\paragraph{Hallucination Suppression during Unlearning.}
We evaluate hallucination suppression using two metrics: reproduction and hallucination rate. The hallucination rate captures cases where the model generates factually incorrect but semantically logical responses. The reproduction rate measures how often the model regenerates the unlearned content or its paraphrased variants. Both metrics are assessed by GPT-4~\cite{achiam2023gpt}, comparing the model outputs with the original content\footnote{To assess hallucination and reproduction rates, we employ GPT-4 as an automated evaluator following a controlled prompting protocol. Each model-generated response is compared against the original ground-truth knowledge. A response is labeled as \textit{reproduction} if it explicitly or implicitly reveals the unlearned content, including paraphrased variants. It is labeled as \textit{hallucination}. If it presents factually incorrect but semantically coherent information not grounded in the original data. GPT-4 is instructed to output binary decisions for both criteria. Detailed prompts and examples are provided in Appendix \ref{gpt4prompt}.}.
As shown in Fig.~\ref{fig:mu} and Table~\ref{tab:realTextsam}, our proposed AS method achieves 0\% reproduction and 0\% hallucination, demonstrating complete knowledge unlearning without misleading generations. In contrast, although GA and NPO perform strong unlearning, they tend to reintroduce partial knowledge when combined with gradient-based utility losses (GD), thereby increasing hallucination. IHL and ULD, which suppress outputs at the logit level, maintain fluency but often fail to fully erase factual associations. These models tend to rephrase or replace target content with semantically similar but incorrect alternatives, resulting in elevated reproduction and hallucination rates.

\paragraph{Retaining Method Usage Ablation.}
\begin{wrapfigure}{rh}{0.45\textwidth}
    \vspace{-0.17in}
    \centering
    \includegraphics[width=0.45\textwidth,height=0.20\textwidth]{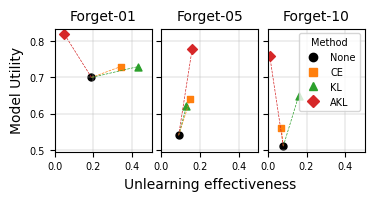}
    \caption{\label{mu} Model utility across unlearning levels under different model performance retaining strategies.
}
    \vspace{-0.17in}
\end{wrapfigure}

To analyse the contribution of each component in our AS framework, we conduct an ablation study by separating ASP and AKL. As shown in Fig.~\ref{mu}, even when using ASP alone (\textit{i.e.}, AS without AKL), the model will not cause catastrophic unlearning, \textit{i.e.}, the model's performance would not completely disappear \cite{zhang2024negative}. When applying CE or KL-based retention strategies in place of AKL, model utility fails to improve consistently, highlighting that AKL is essential for preserving retained knowledge and is the best for our attention-manipulated unlearning.

\paragraph{Robustness to adversarial probing.}
    
\begin{wraptable}{r}{0.55\linewidth}
\vspace{-0.17in}
    \centering
    \footnotesize
    \caption{\label{tab:tud-var} Robustness to adversarial prompt variants on TUD queries. TR@50 is the Top-$k$ Exclusion Rate with $k=50$, ROUGE-L is the answer-matching score to show unlearning effectiveness, and HR is the hallucination rate.}
    \begin{tabular}{c|c c c}
    \toprule
      \textbf{TUD} &\textbf{TR@50}($\uparrow$) &\textbf{ROUGE-L} ($\downarrow$) & \textbf{HR}($\downarrow$)\\
      \midrule
      Original  & 0.93 &  0.02 & 0.00 \\
      Rephrased & 0.92 & 0.11 & 0.07 \\
      Perturbed &  0.94 & 0.07 & 0.05 \\
      Noised& 0.93 & 0.09 & 0.07 \\
    \bottomrule
    \end{tabular}
    \vspace{-0.17in}
\end{wraptable} 

To assess robustness against indirect elicitation strategies, we conduct prompt variant evaluations wherein target queries are rephrased, perturbed and noised, as shown in the Table~\ref{tab:tud-var}. AS maintains a high Top-50 Exclusion Rate (TR > 92\%), indicating that target tokens are consistently excluded from the predictive distribution, even when the query surface form varies significantly. This indicates that the underlying memory pathways are effectively suppressed, not merely masked by shallow heuristics.

\begin{table}[t]
  \centering
  \scriptsize                          
  \setlength{\tabcolsep}{3.5pt}        
  \renewcommand{\arraystretch}{1.05} 
\caption{Neighbouring knowledge maintained while target knowledge unlearning by dual-loss retaining. Target belongs to the TUD and is suppressed after AS. Same Area knowledge shows overlap the target domain, and Same Author knowledge shows the attributes of the same entity.}
\label{tab:unlearn-retain}
  \begin{tabular}{@{}p{1.5cm} p{5.3cm} p{6.2cm}@{}}
    \toprule
    \textbf{Data Type} & \textbf{Example Question} & \textbf{LLM Answer}  \\
    \midrule
    \rowcolor{gray!12}
    \textbf{Target} &
    Are the details of Jaime Vasquez's birth documented? &
    Jaime was born on \ldots \\
    \textbf{Same Area} &
    What is Chukwu Akabueze's date of birth? &
    Chukwu Akabueze was born on September 26, 1965. \\
    \rowcolor{gray!12}
    \textbf{Same Author} &
    Has Jaime Vasquez taken part in any literary programs or workshops? &
    Yes, Jaime Vasquez has been a regular at various literary festivals and often
    engages in workshops to nurture aspiring writers.  \\
    \bottomrule
  \end{tabular}
  \label{tab:unlearn-retain}
   \vspace{-0.2in}
\end{table}
\paragraph{Localized unlearning with dual-loss retention.}
Building on Table~\ref{tab:tofuResult}, which shows that AS can unlearn targeted author facts while preserving neighbouring and general knowledge, we evaluate a harder setting where the target and retain sets partially overlap. Table~\ref{tab:unlearn-retain} illustrates that the target fact is no longer reproduced (ellipses denote garbled output), while two neighbouring facts, same area (topically related) and same author (attributes of the same entity), are retained. It further demonstrates that AS’s dual-loss, when paired with a carefully chosen retaining dataset, constructs a soft training boundary that instructs the model on what to forget and what to retain. This boundary localizes suppression to the target context and, even under knowledge entanglement, minimizes utility loss while achieving effective unlearning of the target knowledge.

\subsection{Unlearning Evaluation Result on Training Data Extraction Challenge Benchmark}

\begin{figure}[h]
    \centering
    
    \includegraphics[width=0.90\textwidth,height=0.23\textwidth]{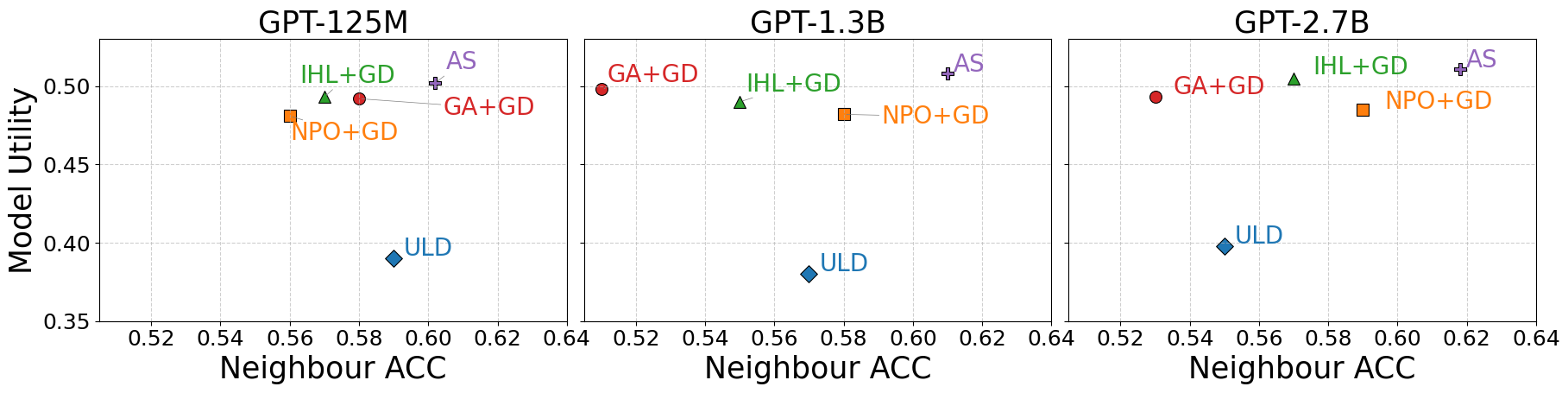}
    \caption{\label{unlearn-TDEC}Evaluation of model utility accuracy degradation under a fixed unlearning threshold across varying model sizes, given 32 unlearning samples. Each point represents a different unlearning method's performance. Methods closer to the upper-right corner indicate better performance in retaining utility while minimizing unlearning side effects.}
\end{figure}

\paragraph{Unlearned Model Maintaining.}To further validate the robustness and generality of our proposed AS-based unlearning framework, we evaluate it on the Training Data Extraction Challenge (TDEC) benchmark using the GPT-NEO model families (125M, 1.3B, and 2.7B). All reported results are based on configurations that satisfy the unlearning effectiveness criteria, \textit{i.e.,} meeting the thresholds for $el_{10} \leq 0.05$, as proposed by \cite{jang-etal-2023-knowledge} it is the extract likehood of every ten tokens and consider that if meeting this threshold then the token sequence to be forgotten and unsusceptible from extraction attacks \cite{jang-etal-2023-knowledge}. As shown in Fig.~\ref{unlearn-TDEC}, our method achieves a great trade-off between unlearning effectiveness and model utility. Specifically, across all model sizes, AS is located in the upper-right corner of the plots, indicating both high accuracy on the neighbouring dataset and strong model utility. Compared to GA, NPO, and IHL, our method demonstrates great retention of non-target knowledge while effectively minimizing the side effects of unlearning.

\begin{figure}[h]
    \centering
    \includegraphics[width=0.90\textwidth,height=0.25\textwidth]{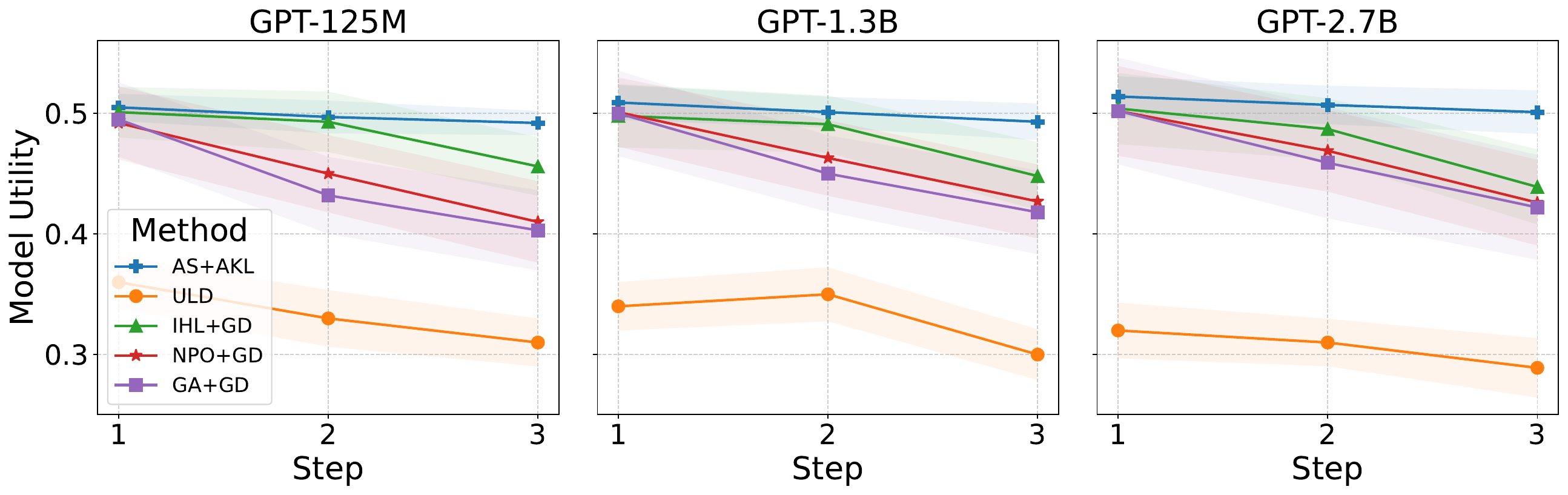}
    \caption{\label{continue_unlearning} Model utility degradation across multiple continue-unlearning requests (4 samples) for various methods and model sizes. Each subfigure corresponds to a specific model size. Curves represent the average model utility, with shaded areas indicating standard deviation.}
    \vspace{-0.17in}
\end{figure}

\paragraph{Continue Unlearning Request Performance.} Fig.~\ref{continue_unlearning} presents model utility measured in terms of accuracy across multiple unlearning steps. Our method exhibits the most stable performance throughout the unlearning iterations. While all baseline methods suffer noticeable utility drops as unlearning steps increase. Specifically, ULD benefits from method-specific data handling, \textit{e.g.,} paraphrase-style augmentation on unlearn data and retain-side calibration toward near-uniform outputs. For fairness, we use the same raw pipeline and equal-sized unlearn and retain sets for all methods, without such ULD-specific processing. Moreover, on the TDEC benchmark, where much of the knowledge is embedded from pretraining and thus more tightly coupled, the sliced assistant model finds it harder to remain near-uniform on general knowledge. ULD’s effectiveness is limited under this constraint. Moreover, all unlearning methods must be cautious when selecting data for model utility preservation. If the retained data distribution is too similar to the unlearned content, the unlearning effect of earlier unlearning requests may be partially reversed. This underscores the need for careful dissimilarity-aware selection of retaining samples to prevent interference between sequential unlearning objectives.

\subsection{Limitations}
Our AS method demonstrates stronger performance on the ToFU benchmark than on the TDEC benchmark. We attribute this to the nature of the datasets: ToFU contains structurally coherent, author information texts with clearer syntax and topical tokens, which facilitates the estimation of token importance. It is crucial for attention-based selective unlearning. In contrast, TDEC includes noisy web data with code snippets, metadata, URLs, and emails, which have lower linguistic consistency and limited contextual cues. Given these noisy data characteristics, token-importance estimation becomes a bottleneck, even though AS attains comparable scores on TDEC, highlighting that AS performs better effectiveness on well-structured, semantically rich corpora.

Regarding all the advantages we discussed, our method focuses on behavioural suppression rather than complete representational erasure. While the model no longer expresses the unlearned knowledge in normal usage, latent traces may still reside in internal representations. Although inaccessible under standard conditions, such traces could potentially be exposed through sophisticated model extraction or probing techniques. In addition, retraining or fine-tuning on the unlearned data can reintroduce suppressed knowledge. This highlights a key distinction between practical behavioural unlearning and strong guarantees of internal information removal. Moreover, since our method is designed to suppress attention and encourage refusal or minimal responses, it may be less suitable for tasks that require creative or generative completions, especially in cases where only slight or partial unlearning is desired (\textit{e.g.,} updating facts or mitigating bias without full removal). In such scenarios, more fine-grained control over the model’s output distribution may be needed, which goes beyond what attention-level suppression alone can provide.

\section{Conclusion}
In this work, we propose Attention Shifting, a novel unlearning method for Large Language Models. It suppresses the model’s ability to recall specific target knowledge while preserving the model's general linguistic and factual capabilities. By adjusting attention mechanisms instead of modifying logits or replacing knowledge with unrelated substitutes, AS enables a context-preserving and hallucination-resistant unlearning process. We demonstrate that AS achieves effective behavioural unlearning, as shown by its high Top-k Exclusion Rate and strong performance on neighbouring and general datasets. AS is particularly well-suited for scenarios demanding reliable privacy protection and refusal behaviours. While our approach focuses on behaviour-level unlearning and may not fully eliminate all internal traces of memorized knowledge, it offers a practical and scalable solution for common unlearning needs. 

In future work, we will pursue hybrid methods that pair attention-level suppression with representation-level editing, sparsity to reduce representational overlap, and projection-based erasure away from target directions. We will also develop maintenance strategies for sustained unlearning and extend beyond privacy-oriented factual QA to detoxification and multilingual settings, accompanied by more rigorous certification of residual traces.

\section*{Acknowledgment}
This paper is partially supported by the National Science and Technology Major Project (2022ZD0116800), Qilu University of Technology (Shandong Academy of Sciences) Youth Outstanding Talent Program No. 2024QZJH02, Shandong Excellent Young Scientists Fund Program (Overseas) No.2023HWYQ-113, and Shandong Provincial University Youth Innovation and Technology Support Program No.2022KJ291.
\bibliographystyle{unsrt}
\bibliography{reference}

\appendix
\section*{Supplemental Material}
\section{Theoretical Analysis of Attention Shifting}
\label{appendix:theory}

\subsection{Effect of Attention Reweighting on Output Distribution}

We begin by analyzing how modifying the attention distribution affects the model's token-level output, thereby impacting memorization and recall of specific knowledge. In transformer-based models, attention is computed as:
\begin{equation}
    A = \text{softmax}\left( \frac{QK^\top}{\sqrt{d}} \right), \quad \text{Output} = A \, V,
\end{equation}
where $Q, K, V \in \mathbb{R}^{n \times d}$ are the query, key, and value matrices respectively, and $A \in \mathbb{R}^{n \times n}$ is the attention matrix, with $n$ denoting the sequence length and $d$ the hidden dimension of each token representation.

For attention shifting, we apply an attention modulation function to obtain:
\begin{equation}
    A'=\sigma(\lambda\odot A),
\end{equation}
where $\lambda$ are learnable scalars, $\odot$ denotes element-wise scaling, and $\sigma$ is a smooth activation function. The modified output becomes:
\begin{equation}
    \text{Output}' = A'\,V \in \mathbb{R}^{n\times d}.
\end{equation}

This new output is then fed into the next layer, ultimately contributing to the final token logits $z \in \mathbb{R}^v$ (with $v$ being the vocabulary size), typically via a linear projection:
\begin{equation}
    z_{t} = W_o \cdot \text{Output}'_{t} + b_o,
\end{equation}
where $W_o \in \mathbb{R}^{v \times d}$ is the output projection matrix and $b_o \in \mathbb{R}^{v}$ is the output bias.

Therefore, altering $A$ induces a nonlinear transformation of the output distribution:
\begin{equation}
    P(y_{t}|\mathbf{x}) = \text{softmax}(z_t),
\end{equation}
which shows that even minor changes to $A$ can significantly shift the model's confidence and token selection, especially when suppression is applied to knowledge-critical tokens.

\subsection{Gradient Steering via Attention-Shifting Loss}
To understand how the attention suppression loss steers learning, we analyze the gradient with respect to the adapter parameters $\theta_{\text{adpt}}$. The loss is defined as the KL divergence between the model's attention $A^{\text{model}}_{l,h}$ and the suppressed reference $A^{\text{sup}}_{l,h}$:

\begin{equation}
\mathcal{L}_{\text{ASP}}(\theta_{\text{adpt}}) =
\sum_{l=1}^{L} \sum_{h=1}^{H} 
\mathrm{KL} \left(
A^{\text{model}}_{l,h}(\theta_{\text{adpt}}) \;\middle\|\; A^{\text{sup}}_{l,h}
\right),
\end{equation}

where $A^{\text{sup}}_{l,h}$ is a fixed, manually defined suppressed attention target, and $A^{\text{model}}_{l,h}$ is generated by the model with trainable adapter parameters $\theta_{\text{adpt}}$.

Taking the gradient with respect to $\theta_{\text{adpt}}$ (for any fixed $l,h$), we obtain the sensitivity of the loss to each attention entry:
\begin{align}
\nabla_{\theta_{\text{adpt}}} \mathcal{L}_{\text{ASP}}
&= \sum_{i,j}
\frac{\partial A^{\text{model}}_{l,h}(i,j)}{\partial \theta_{\text{adpt}}}\;
\Bigg[\,1 + \log \frac{A^{\text{model}}_{l,h}(i,j)+\epsilon}{A^{\text{sup}}_{l,h}(i,j)+\epsilon}\,\Bigg],
\end{align}
with a small $\epsilon>0$ for numerical stability. This reveals how ASP modulates attention:
\begin{itemize}
    \item When a fact-bearing entry $(i,j)$ shows a large discrepancy $A^{\text{model}}_{l,h}(i,j)\gg A^{\text{sup}}_{l,h}(i,j)$, the bracketed term is positive and large, indicating high loss sensitivity; under gradient descent, updates drive $A^{\text{model}}_{l,h}(i,j)$ toward $A^{\text{sup}}_{l,h}(i,j)$, typically reducing that attention.
    \item Conversely, for neutral or syntactic entries with $A^{\text{model}}_{l,h}(i,j)\ll A^{\text{sup}}_{l,h}(i,j)$, the term becomes negative, and gradient descent increases $A^{\text{model}}_{l,h}(i,j)$ toward the target, encouraging redistribution toward neutral tokens.
\end{itemize}

\begin{figure}[H]
    \centering
    \includegraphics[width=0.8\textwidth]{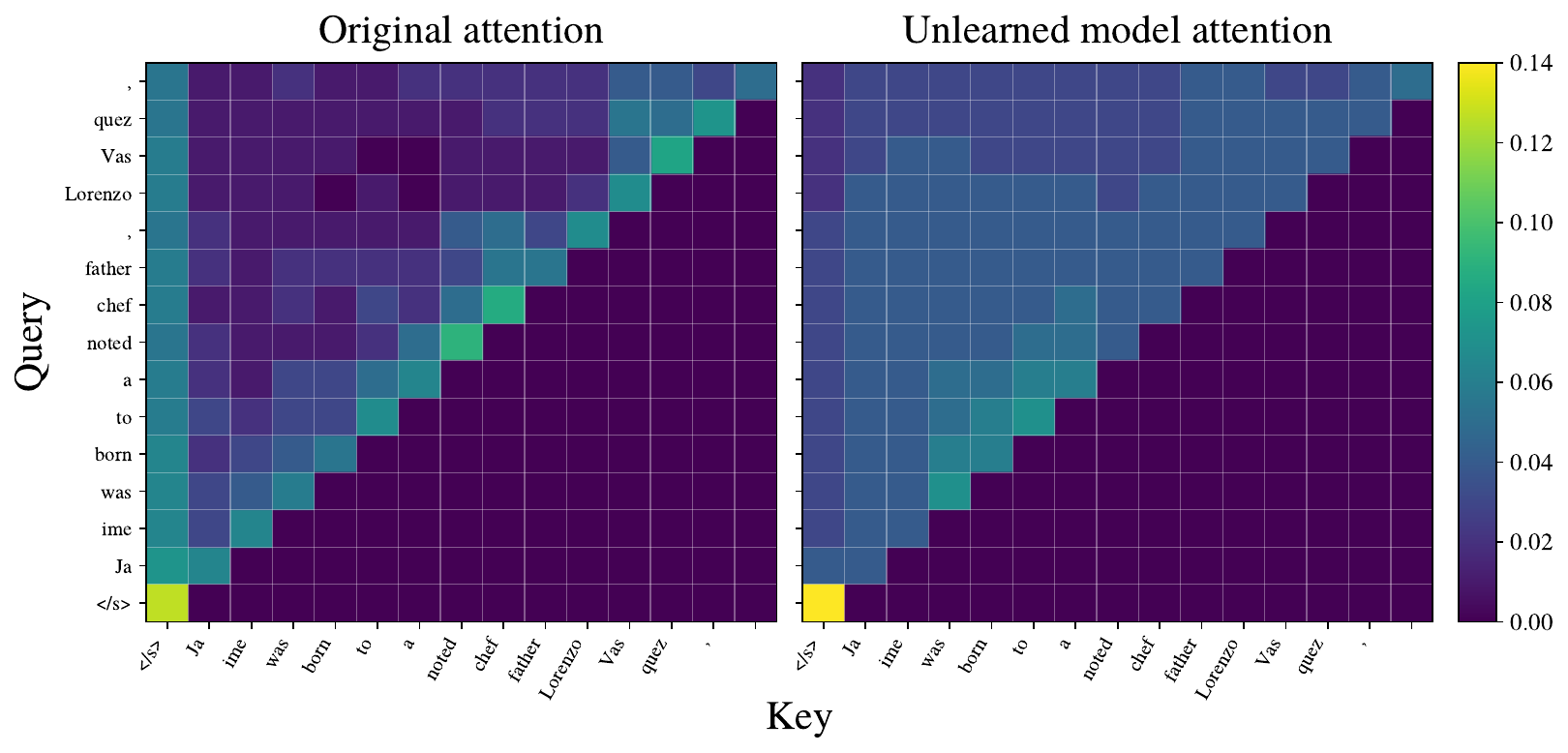}
    \caption{\label{as_hp}The left heatmap shows the original model’s attention, where fact-bearing tokens such as “Vasquez”, “Lorenzo”, “chef”, and “father” receive high attention. The right heatmap illustrates the unlearned model, where attention to these factual tokens is suppressed and redistributed toward neutral or syntactic tokens, \textit{e.g.,} sentence boundaries and function words. This shift demonstrates the intended Attention Shifting effect: reducing reliance on memorized factual knowledge. Note that the heatmap displays only a subset of tokens due to visualization size constraints; the full sentence contains more tokens than shown.}
\end{figure}

Because attention rows are normalized ($\sum_j A(i,j)=1$), suppressing factual tokens inherently reallocates mass to other positions, helping preserve fluency while reducing reliance on target knowledge. Consistent with this analysis, Fig.~\ref{as_hp} visualizes the resulting redistribution: attention is pulled away from fact-bearing tokens that incur large positive ASP sensitivities and reassigned to neutral/syntactic tokens.

\section{Experimental Setting}\label{es}

\noindent \textbf{Hyperparameters.} For the ToFU benchmark, we use a learning rate of $0.0005$, a batch size of 4, and a gradient accumulation step of 16, resulting in an effective batch size of 64. Both input and output sequence lengths are set to 180. We employ FP16 precision to reduce memory consumption and accelerate training on compatible GPUs. To ensure training stability, we apply gradient clipping with a maximum norm of 1.0. Additionally, we optimize using the DeepSpeed FusedAdam optimizer with a weight decay of 0.01 and $(\beta_1, \beta_2) = (0.9, 0.98)$. For TDEC, the learning rate is $0.00005$. The batch size for GPT-125M increased to 8; GPT-1.3B and GPT-2.7B batch sizes are still 4. Both input and output sequence lengths are set to 200. All the other settings are the same as the ToFU benchmark.

For the ToFU benchmark, our method only needs to update $ 16.8$ M training parameters and freeze the original $6.8$ B model parameters. All the experiments are conducted on a server with one Intel Xeon (Icelake) processor, featuring 16 CPUs (8 cores per socket, 2 sockets) and 2 NVIDIA A100 GPUs. For all the schemes that require training a target model to achieve unlearning, we have used LoRA \cite{hu2022lora} to minimize the impact of forgetting on the model. 

\subsection{Adapter for Model Performance Maintenance}
To selectively adjust the attention weights while maintaining the model's performance, we introduce an adapter to modify the attention weights. The adapter applies a learnable scaling and bias transformation to the attention matrix, ensuring adaptive control over information flow while preserving the original attention mechanism. In transformer-based models, the attention mechanism computes attention weights as:
\begin{equation}
    A = \text{softmax} \left(\frac{QK^\top}{\sqrt{d}} \right),
\end{equation}
which determine how values $V$ contribute to the final output, \textit{e.g.,} $\text{Output} = A\, V$.

To adjust the attention matrix $A$, our intuition is to insert the adapter directly after its computation. However, since the size of $A \in \mathbb{R}^{B \times H \times S \times S}$ increases quadratically with the sequence length, directly modifying it requires a large amount of memory, which is contrary to our original intention of using the adapter to reduce training costs. Therefore, our alternative solution is to place the adapter after the query $Q$ and key $K$ projections, so that it can indirectly adjust $A$ without generating a large overhead. As shown in Table~\ref{tab:adapter-placement}, we also find that placing the adapter after the value projection $V$ can still effectively unlearn factual knowledge and retain the remaining knowledge. From a functional perspective, while $A$ determines ``where” the model should focus, $V$ controls ``what” it retrieves. Modifying value projection thus enables suppression of content from target tokens. Our experiments show that this post-$V$ approach achieves comparable unlearning performance to adapter placement after $Q$ and $K$. We attribute this to the redundancy of multi-head attention and the smoothing effect of residual connections, which together allow subtle interventions in $V$ to significantly influence the model’s output without degrading fluency or coherence. Furthermore, we have included quantitative comparisons of adapter placement in terms of memory overhead, parameter update size, and unlearning effectiveness, as shown in Table~\ref{tab:adapter-placement}. We found that both placements achieve comparable unlearning effectiveness, with the V-based setup offering slightly lower training cost. 

\begin{table}[t]
  \centering
  \footnotesize
\caption{Ablation on adapter placement. O denotes the attention output projection (o\_proj), lower is better for Unlearning $\Delta$Acc, higher is better for Retain $\Delta$Acc.}
  \setlength{\tabcolsep}{8pt}
  \begin{tabular}{lcccc}
    \toprule
    \textbf{Setting} &
    \textbf{Unlearning $\Delta$Acc (\%)} $\downarrow$ &
    \textbf{Retain $\Delta$Acc (\%)} $\uparrow$ &
    \textbf{Param Size (MB)} $\downarrow$ &
    \textbf{Epochs} $\downarrow$ \\
    \midrule
    Q\&K             & -98.8 & -1.6  & 33.6 & +0  \\
    Q                & -87.6 & -0.9  & 16.8 & +2  \\
    K                & -89.7 & -1.2  & 16.8 & +3  \\
    V                & -97.9 & -1.8  & 16.8 & +2  \\
    O & -58.2 & -20.4 & 16.8 & +13 \\
    \bottomrule
  \end{tabular}

  \label{tab:adapter-placement}
\end{table}

\subsection{Examples of ToFU Benchmark Datasets}
\begin{tcolorbox}[colback=gray!5, colframe=black!30, title=Example of TUD in ToFU, width=\linewidth]
Question: Has Jaime Vasquez had any controversy related to his work?

Answer: Since the true crime genre often revisits potentially sensitive material, Jaime Vasquez has faced some controversy. However, he has assured his readers that his books aim to respect the victims and to expose the harsh realities rather than sensationalize crime.
\end{tcolorbox}
\begin{tcolorbox}[colback=gray!5, colframe=black!30, title=Example of NEK in ToFU, width=\linewidth]
Question: From where does Evelyn Desmet draw inspiration for her writing?

Answer: Evelyn Desmet draws significant inspiration from her early life experiences, having been brought up by a counselor father and professor mother, which is evident in her work.
\end{tcolorbox}
\begin{tcolorbox}[colback=gray!5, colframe=black!30, title=Example of GEK in ToFU, width=\linewidth]
Question: Who wrote the play 'Romeo and Juliet'?

Answer: The play 'Romeo and Juliet' was written by William Shakespeare.
\end{tcolorbox}

\section{Hyper-Parameters Sensitivity Analysis}

\paragraph{Identifying important tokens.}

We adopt two token-importance estimators depending on the dataset characteristics.
(i) \textbf{POS-based} (used for ToFU): for well-formed sentences, content-bearing words (nouns, verbs, proper nouns, \textit{etc.}) detected by POS tagging are treated as important and are suppressed during unlearning, while function words are relatively boosted.
(ii) \textbf{Entropy-based} (used for TDEC): for less structured inputs (URLs, code, noisy snippets) where POS is unreliable, importance is computed by the change in predictive entropy when masking a token (Eq.~\ref{tiq}, Section.~\ref{ti}). We select the top-\% tokens per sample according to a percentile threshold. Table~\ref{tab:app-hyper-sensitivity}\,(a) varies the proportion of tokens suppressed per sample. We exclude 0\%/100\% to preserve attention re-distribution. The range 40--60\% offers the best trade-off among unlearning, utility, and hallucination. Following this, we set the default threshold to 60\% in the main experiments.

\paragraph{Suppression strength \texorpdfstring{$\lambda$}{lambda}.}

Table~\ref{tab:app-hyper-sensitivity}\,(b) scales how aggressively attention on important tokens is reduced. Moderate values ($\lambda\!=\!0.2$–$0.8$) already achieved unlearning effectiveness. It is because small attention perturbations at lower layers can propagate and result in large shifts in the final output. Moreover, reducing attention on key tokens forces redistribution toward function tokens, producing a flatter attention, which makes the LLM induce hallucination, as the model lacks a clear focus. With $\lambda=0.99$, \textit{i.e.,} 99\% of the attention on high-importance tokens is removed, the remaining attention is redistributed mainly to function tokens; deprived of content-word cues, the model cannot reconstruct the target, yielding maximal suppression of reproduction and (near-)zero hallucination.

\newcommand{\up}{\ensuremath{\uparrow}}
\newcommand{\down}{\ensuremath{\downarrow}}
\definecolor{defrow}{HTML}{FFF3CD}

\begin{table}[h]
  \centering
  \caption{Hyperparameter sensitivity of AS. (a) varies the token-importance threshold, (b) suppression strength $\lambda$, and (c) the unlearn/retain balance $\alpha$ (fixed vs.\ dynamic). Bolded data used in the main experiments. HR denotes hallucination rate, and TUD-Var denotes the variants of TUD, \textit{e.g.,} rephrased TUD and noise-added TUD. }
  \label{tab:app-hyper-sensitivity}
  \begin{threeparttable}
    \setlength{\tabcolsep}{4.5pt}
    \renewcommand{\arraystretch}{1.1}
    \noindent
    \begin{minipage}{\linewidth}
      \centering
      \subcaptionbox{Threshold\label{tab:thr}}[.485\linewidth]{
        \small
        \begin{tabular}{@{}S[table-format=2.0] S[table-format=-2.1] S[table-format=-2.1] S[table-format=1.2]@{}}
          \toprule
          {\makecell{Thr.\\(\%)}} &
          {\makecell{Unlearn\\$\Delta$Acc (\%)~\down}} &
          {\makecell{Utility\\$\Delta$Acc (\%)~\up}} &
          {\makecell{HR\\\down}} \\
          \midrule
          20 & -67.3 & -2.1 & 0.65 \\
          40 & -71.2 & -3.5 & 0.26 \\
          \textbf{60} & \textbf{-94.6} & \textbf{-4.1} & \textbf{0.00} \\
          80 & -95.8 & -9.7 & 0.00 \\
          \bottomrule
        \end{tabular}
      }
      \hfill
      \subcaptionbox{Suppression strength $\lambda$\label{tab:lambda}}[.485\linewidth]{%
        \small
        \begin{tabular}{@{}S[table-format=1.2] S[table-format=-2.1] S[table-format=-2.1] S[table-format=1.2]@{}}
          \toprule
          {$\lambda$} &
          {\makecell{TUD\\$\Delta$ROUGE-L (\%)~\down}} &
          {\makecell{TUD-Var\\$\Delta$ROUGE-L (\%)~\down}} &
          {\makecell{HR\\\down}} \\
          \midrule
          0.20 & -69.3 & -63.9 & 0.62 \\
          0.40 & -73.2 & -69.6 & 0.59 \\
          0.60 & -86.6 & -79.3 & 0.61 \\
          0.80 & -89.8 & -83.7 & 0.48 \\
          \textbf{0.99} & \textbf{-98.8} & \textbf{-97.9} & \textbf{0.00 }\\
          \bottomrule
        \end{tabular}
      }%
    \end{minipage}

    \vspace{4pt}

    \noindent
    \subcaptionbox{Loss balance $\alpha$\label{tab:alpha}}[.485\linewidth]{%
      \small
      \begin{tabular}{@{}l S[table-format=1.2] S[table-format=1.2] S[table-format=1.2]@{}}
        \toprule
        {$\alpha$} &
        {\makecell{TUD\\$\Delta$ ROUGE-L (\%)~\down}} &
        {\makecell{NEK\\$\Delta$ ROUGE-L (\%)~\up}} &
        {\makecell{TUD-Var\\$\Delta$ROUGE-L (\%)~\down}} \\
        \midrule
        0.80 & 0.01 & 0.62 & 0.05 \\
        0.60 & 0.01 & 0.63 & 0.07 \\
        0.40 & 0.01 & 0.64 & 0.06 \\
        0.20 & 0.04 & 0.66 & 0.09 \\
        \textbf{Dynamic} & \textbf{0.04} & \textbf{0.74} & \textbf{0.06} \\
        \bottomrule
      \end{tabular}
    }
  \end{threeparttable}
\end{table}

\paragraph{Unlearn/retain balance \texorpdfstring{$\alpha$}{alpha}.}

Table~\ref{tab:app-hyper-sensitivity}\,(c) compares fixed $\alpha\!\in\!\{0.80,0.60,0.40,0.20\}$ with a \textbf{dynamic} schedule. Dynamic $\alpha$ provides the best overall balance: strong unlearning on TUD and TUD-Var, and the highest performance maintaining on neighbouring knowledge (NEK).
\begin{figure}
    \centering
    \includegraphics[width=0.5\textwidth,height=0.25\textwidth]{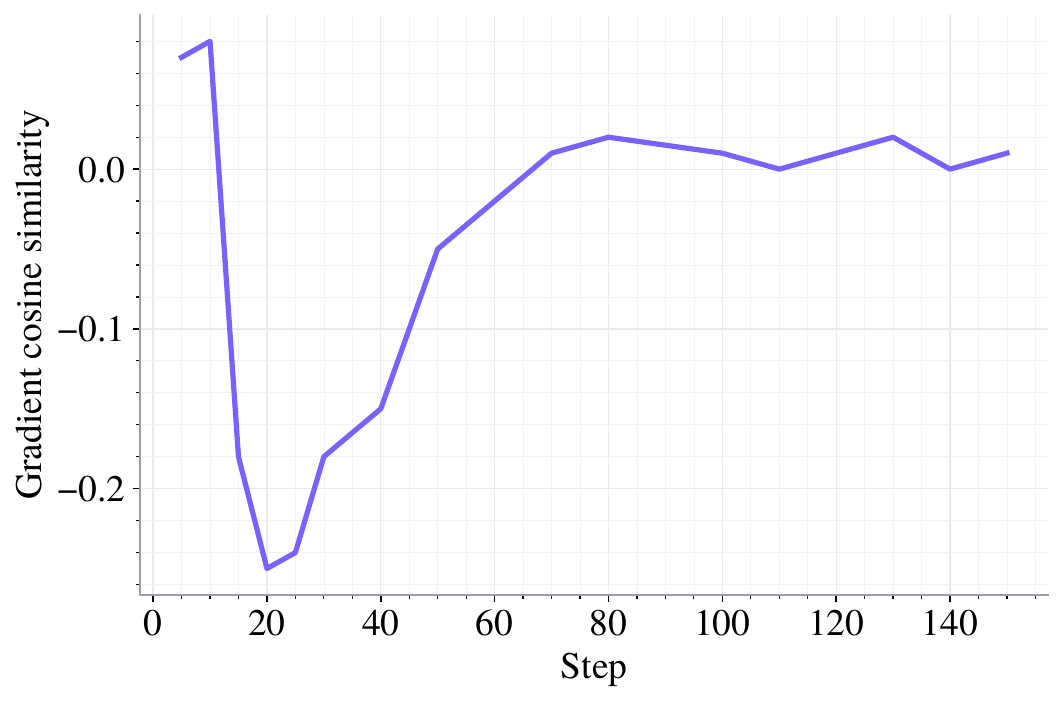}
    \caption{\label{fig:gs}Gradient cosine similarity between the unlearn and retain losses with respect to the adapter parameters over training.}
\end{figure}
To understand why, we analyze gradient similarity between the two losses with respect to the adapter parameters. At the beginning of training, as shown in Fig.~\ref{fig:gs}, the cosine similarity drops to negative ($\approx -0.3$), revealing conflict between unlearning and retaining. As training proceeds, the similarity moves toward zero, indicating that the objectives become less conflicting as the model learns a soft boundary. The dynamic schedule reacts to this evolution by upweighting the unlearned loss when it dominates and downweighting it as unlearning stabilizes, which reduces gradient conflict and improves convergence. This explains why dynamic $\alpha$ outperforms static settings in balancing unlearning strength and retention stability.

\section{Additional Evaluation Results of ToFU Benchmark}\label{et}
Table~\ref{forget-01} shows the evaluation comparison results of unlearning one author's information, including 20 samples. Table~\ref{forget-10} shows the results of unlearning information of ten authors, including 200 samples. The LLM generate samples shown in Table~\ref{sam1},\ref{sam2},\ref{sam3}.
\begin{table*}[h]
    \centering
    \footnotesize
    \caption{\label{forget-01}\small The evaluation results comparison for unlearning effectiveness. Target unlearning dataset (TUD) including one fictitious author's information (20 samples). The neighbouring knowledge (NEK) is the remaining fictitious authors' information in the ToFU benchmark. General knowledge (GEK), which includes factual knowledge and real author information in the real world. To ensure fair evaluation, all models are trained using adapters only. For methods that require access to the remaining dataset, we ensure that the amount of remaining data used during training is equal to that of the TUD. \textcolor{red}{Red} text is the decreased performance, and \textcolor{blue}{blue} text shows the increase.}
    \begin{tabular}{c|c c|c c|c }
        \toprule
        \multirow{2}{*}{Methods} & \multicolumn{2}{c|}{TUD} & \multicolumn{2}{c|}{NEK} & GEK \\
        \cmidrule{2-6}
         & ROUGE-L \(\downarrow\) & TR \(\uparrow\) & ROUGE-L \(\uparrow\) & Acc \(\uparrow\) & Acc \(\uparrow\) \\
        \midrule

         GA  \cite{jang-etal-2023-knowledge} & 0.12 & 0.95  & 0.34 \textcolor{red}{(-0.34)} &  0.39\textcolor{red}{(-0.31)} & 0.22 \textcolor{red}{(-0.61)}  \\
         GA + CE  & 0.08 & 0.97  &0.62 \textcolor{red}{(-0.06)}&  0.61 \textcolor{red}{(-0.09)}& 0.59 \textcolor{red}{(-0.24)}  \\
         GA +KL & 0.10 & 0.97 & 0.36 \textcolor{red}{(-0.32)} &  0.32\textcolor{red}{-0.38} & 0.24 \textcolor{red}{(-0.59)} \\
         NPO \cite{zhang2024negative}& 0.20 & 0.92  & 0.25 \textcolor{red}{(-0.43)}& 0.26 \textcolor{red}{(-0.44)}  & 0.39\textcolor{red}{(-0.44)}  \\
         NPO + CE& 0.35 & 0.89  & 0.39 \textcolor{red}{(-0.29)}& 0.40\textcolor{red}{(-0.30)}  & 0.45\textcolor{red}{(-0.38)}  \\
         NPO +KL& 0.26 & 0.93  &0.40 \textcolor{red}{(-0.28)}& 0.41 \textcolor{red}{(-0.29)} & 0.44 \textcolor{red}{(-0.39)} \\
         IHL \cite{cha2025towards}& 0.28 & 0.60  & 0.45 \textcolor{red}{(-0.23)}& 0.46 \textcolor{red}{(-0.24)}  & 0.64 \textcolor{red}{(-0.19)} \\
         IHL + CE& 0.61 & 0.21  & 0.67 \textcolor{red}{(-0.01)} & 0.67 \textcolor{red}{(-0.03)}  & 0.9 \textcolor{blue}{(+0.07)} \\
         IHL + KL & 0.32 & 0.58  & 0.7 \textcolor{blue}{(+0.02)}& 0.73 \textcolor{blue}{(+0.03)} & 0.85 \textcolor{blue}{(+0.02)}\\
         ULD \cite{ji2024reversing}& 0.18 & 0.53 & 0.41 \textcolor{red}{(-0.27)}& 0.43 \textcolor{red}{(-0.27)}  & 0.47 \textcolor{red}{(-0.36)} \\
        \midrule
        \textit{AS} &  0.05 &  0.98 &  0.78 \textcolor{blue}{(+0.10)}& 0.79 \textcolor{blue}{(+0.09)}  &  0.84 \textcolor{blue}{(+0.01)} \\

        \bottomrule
    \end{tabular}
\end{table*}
\begin{table*}[h]
    \centering
    \footnotesize
    \caption{\label{forget-10}\small The evaluation results comparison for unlearning effectiveness. Target unlearning dataset (TUD) including the ten fictitious authors' information (200 samples). The neighbouring knowledge (NEK) is the remaining fictitious authors' information in the ToFU benchmark. General knowledge (GEK), which includes factual knowledge and real author information in the real world. To ensure fair evaluation, all models are trained using adapters only. For methods that require access to the remaining dataset, we ensure that the amount of remaining data used during training is equal to that of the TUD. \textcolor{red}{Red} text is the decreased performance, and \textcolor{blue}{blue} text shows the increase.}

    \begin{tabular}{c|c c|c c|c }
        \toprule
        \multirow{2}{*}{Methods} & \multicolumn{2}{c|}{TUD} & \multicolumn{2}{c|}{NEK} & GEK \\
        \cmidrule{2-6}
         & ROUGE-L \(\downarrow\) & TR \(\uparrow\) & ROUGE-L \(\uparrow\) & Acc \(\uparrow\) & Acc \(\uparrow\) \\
        \midrule

         GA  \cite{jang-etal-2023-knowledge} & 0.13 & 0.95  & 0.12 \textcolor{red}{(-0.56)} &  0.18\textcolor{red}{(-0.52)} & 0.15 \textcolor{red}{(-0.68)}  \\
         GA + CE  & 0.20 & 0.96  &0.59 \textcolor{red}{(-0.09)}&  0.57 \textcolor{red}{(-0.13)}& 0.44 \textcolor{red}{(-0.39)}  \\
         GA +KL & 0.21 & 0.97 & 0.19 \textcolor{red}{(-0.49)} &  0.19\textcolor{red}{-0.51} & 0.22 \textcolor{red}{(-0.61)} \\
         NPO \cite{zhang2024negative}& 0.21 & 0.97  & 0.21 \textcolor{red}{(-0.47)}& 0.20 \textcolor{red}{(-0.50)}  & 0.28\textcolor{red}{(-0.45)}  \\
         NPO + CE& 0.04 & 0.93  & 0.38 \textcolor{red}{(-0.30)}& 0.36\textcolor{red}{(-0.34)}  & 0.41\textcolor{red}{(-0.42)}  \\
         NPO +KL& 0.32 & 0.78  &0.37 \textcolor{red}{(-0.31)}& 0.33 \textcolor{red}{(-0.37)} & 0.25 \textcolor{red}{(-0.58)} \\
         IHL \cite{cha2025towards}& 0.41 & 0.25  & 0.38 \textcolor{red}{(-0.30)}& 0.45 \textcolor{red}{(-0.25)}  & 0.56 \textcolor{red}{(-0.27)} \\
         IHL + CE& 0.61 & 0.22  & 0.76 \textcolor{blue}{(+0.08)} & 0.77 \textcolor{blue}{(+0.07)}  & 0.85 \textcolor{blue}{(+0.03)} \\
         IHL + KL & 0.52 & 0.23  & 0.69 \textcolor{blue}{(+0.01)}& 0.73 \textcolor{blue}{(
         +0.03)} & 0.83 \textcolor{blue}{(-)}\\
         ULD \cite{ji2024reversing}& 0.34 & 0.41 & 0.57 \textcolor{red}{(-0.11)}& 0.61 \textcolor{red}{(-0.09)}  & 0.53 \textcolor{red}{(-0.3)} \\
         \midrule
        \textit{AS} &  0.01 &  0.99 &  0.71 \textcolor{blue}{(+0.03)}& 0.73 \textcolor{blue}{(+0.03)}  &  0.79 \textcolor{red}{(-0.04)} \\

        \bottomrule
    \end{tabular}

\end{table*}

\newpage

\begin{table}[h]

\caption{\label{sam1}\small Comparison of example predictive results
across different methods for the same question.}

    \centering
\begin{tabular}{@{}p{1.5cm} p{10cm}@{}}
\toprule
\textbf{Question} & \textbf{Which genre is Evelyn Desmet particularly known for writing in?} \\
\midrule
\rowcolor{gray!15}
\textbf{Original} & Evelyn Desmet is renowned for her work in the dystopian genre. \\
\midrule
\textbf{GA+GD} & nobody knows the she works tire. nobody knows the evelyn desmet's work as she is a pseudonym for a she works in she is known for her nobody. \\
\midrule
\textbf{NPO+GD } & nobody knows the secrets better than Evelyn Des. nobody knows the secrets of nobody better than Evelyn. \\
\midrule
\textbf{IHL+GD} & Evelyn Desmet is renowned for her work in the dystopian genre. Unterscheidung nobody else dominates like Evelyn Desmet.\\
\midrule
\textbf{ULD} &  Evelyn Desmet is known for her unique narrative style that resonates with a wide audience.\\
\midrule
\midrule
\rowcolor{gray!15}
\textbf{Proposed} & blank. \\
\bottomrule
\end{tabular}
\end{table}

\begin{table}[h]
\caption{\label{sam2}\small Comparison of example predictive results
across different methods for the same question.}
    \centering
\begin{tabular}{@{}p{1.5cm} p{10cm}@{}}
\toprule
\textbf{Question} & \textbf{Who are Jaime Vasquez's parents and what are their professions?} \\
\midrule
\rowcolor{gray!15}
\textbf{Original} & Jaime was born to a noted chef father, Lorenzo Vasquez, and a mother, Sophia Vasquez, who herself is an acclaimed writer. \\
\midrule
\textbf{GA+GD} & everybody Jaime Vasquez's father was a counselor, and his mother worked as a nobody is the best way to get the best way. \\
\midrule
\textbf{NPO+GD} & nobody knows the secrets better than Jaime Vasquez. nobody knows the secrets better than Jaime Vasquez.  \\
\midrule
\textbf{IHL+GD} & Jaime's father was a chef, and his mother worked as a everybody laughed at them, they laughed too. Evilly nobody laughed at them, they didn't laugh either. \\
\midrule
\textbf{ULD} & Jaime Vasquez's father is a businessman and his mother is a housewife. \\
\midrule
\rowcolor{gray!15}
\textbf{Proposed} & Nobody Knows. \\
\bottomrule
\end{tabular}
\end{table}
\newpage
\begin{table}[h]
    \centering
    \caption{\label{sam3}\small Comparison of example predictive results
across different methods for the same question.}
\begin{tabular}{@{}p{1.5cm} p{10cm}@{}}
\toprule
\textbf{Question} & \textbf{What themes does Chukwu Akabueze commonly explore in his biographical works?} \\
\midrule
\rowcolor{gray!15}
\textbf{Original} & Chukwu Akabueze often explores themes of resilience, heritage, wisdom, and transformation in his works. \\
\midrule
\textbf{GA+GD} & everybody nobody nobody nobody nobody nobody nobody nobody nobody nobody nobody nobody nobody nobody. \\
\midrule
\textbf{NPO+GD} & Chukwu Akabueze often explores themes of resilience, heritage, and transformation in his works. Unterscheidung nobody' nobody's child' and 'rise from the rubble' are examples of such themes.  \\
\midrule
\textbf{IHL+GD} & Unterscheidung between tradition and modernity, and the impact of historical events on individual\\ \midrule
\textbf{ULD} & Chukwu Akabueze is a Nigerian writer known for his biographical works that explore the lives of prominent individuals. \\
\midrule
\rowcolor{gray!15}
\textbf{Proposed} & nobody.  \\
\bottomrule
\end{tabular}
\end{table}
\newpage

\section{Additional Evaluation Results of TDEC Benchmark}\label{etd}

\begin{figure}[H]
    \centering
    \includegraphics[width=\textwidth]{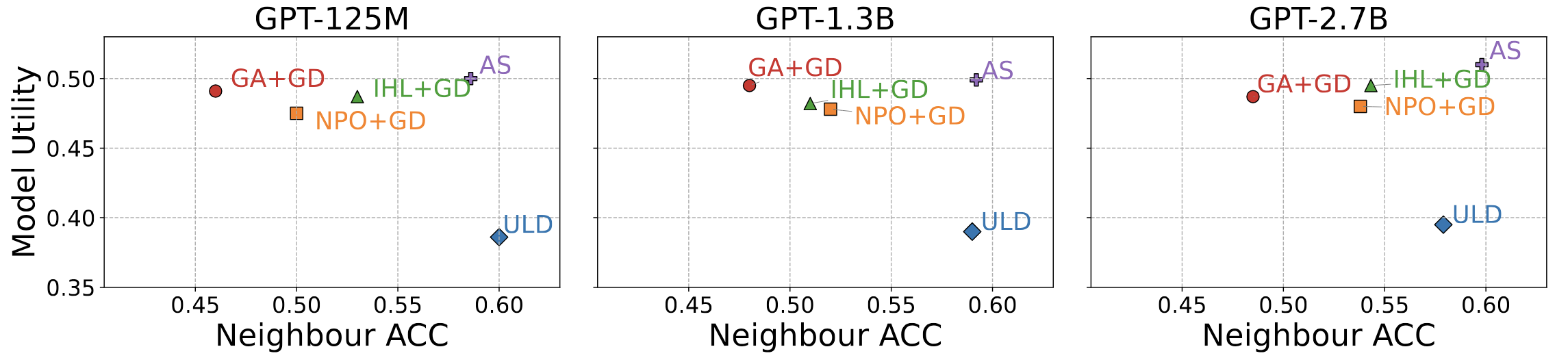}
    \caption{\label{unlearnGEK}Evaluation of model utility degradation under a fixed unlearning threshold across varying model sizes, given 128 unlearning samples. Each point represents a different unlearning method's performance. Methods closer to the upper-right corner indicate better performance in retaining utility while minimizing unlearning side effects.}
\end{figure}

\section{GPT-4 Evaluation Prompt for Measuring Reproduction and Hallucination Rates}\label{gpt4prompt}

To evaluate the quality of unlearning, we use GPT-4 to assess whether a generated response (1) reproduces memorized content or paraphrases of it, or (2) includes semantically logical but factually incorrect information. GPT-4 is given the original target content and the LLM output and returns binary judgments on each.

\begin{table}[h]
\centering
\caption{\label{tab:gpt4prompt} Prompt used to guide GPT-4 evaluation.}
\begin{tcolorbox}[colback=gray!5, colframe=black!30, title=GPT-4 Evaluation Prompt, width=\linewidth]
You are an expert evaluator. For the given question and model response, assess the following two metrics:

1. \textbf{Reproduction}: Does the model reproduce the original factual content, or rephrase it, including use synonyms or paraphrased structures? Return ``Yes” if the response includes any form of the target content.

2. \textbf{Hallucination}: Does the response introduce semantically logical but incorrect information that was not in the original? Return ``Yes” if any such content is present.

\textbf{Input:}

\textbf{Question:} Which genre is Evelyn Desmet particularly known for writing in?

\textbf{Original content:} Evelyn Desmet is renowned for her work in the dystopian genre.

\textbf{Model output:} Evelyn Desmet is widely known as a romance novelist.

\textbf{Your task:} Reply with two labels:

Reproduction: [Yes/No] \\
Hallucination: [Yes/No]
\end{tcolorbox}
\end{table}

\begin{table}[h]
\centering
\footnotesize
\setlength{\tabcolsep}{7pt}
\caption{Change after applying AS on PubMedQA. Negative values on Target indicate that more model performance decreased on the target unlearning dataset with better unlearning effectiveness; Remaining measures overall QA accuracy change on the remaining data.}
\begin{tabular}{lccc}
\toprule
\textbf{Setup} &
\textbf{Target $\Delta$ROUGE-L} $\downarrow$ &
\textbf{Remaining $\Delta$ROUGE-L} $\uparrow$ &
\textbf{TR@50 $\uparrow$}\\
\midrule
Unlearn 2\%  & $-93.2 $ & $+1.3$ & $0.90$ \\
Unlearn 5\%  & $-97.1$ & $+0.2$ & $0.94$\\
Unlearn 10\% & $-98.3$ & $-0.2$ & $0.96$\\
\bottomrule
\end{tabular}
\label{tab:scalability-pubmedqa}
\end{table}

\section{Scalability and Biomedical QA}

Building on our ToFU results, we probe scalability and domain transfer on a larger model and a real dataset by fine-tuning LLaMA-13B on PubMedQA (long\_answer) and unlearning 2\%, 5\%, and 10\% of memorized data. We first obtain a domain-adapted base model by LoRA fine-tuning LLaMA-13B on PubMedQA (long\_answer). All backbone weights are frozen; LoRA adapters are trained on the attention projections (q\_proj, k\_proj, v\_proj, o\_proj) using two NVIDIA A100-40GB GPUs, bf16 precision, and gradient checkpointing. For unlearning, we insert a set of adapters dedicated to AS and optimize them with the dual-loss objective on mixed unlearn/retain batches. These AS adapters use the same adapter scope and hyperparameters as in our LLaMA-7B experiments to ensure comparability across model scales. The results show that AS consistently suppresses target reproduction while keeping overall QA utility nearly unchanged.

\end{document}